\documentclass[sn-nature,pdflatex,iicol]{sn-jnl}
\pdfoutput=1

\usepackage{multirow}%
\usepackage{bbm} 
\usepackage{manyfoot}%
\usepackage{booktabs}%
\usepackage{algorithm}%
\usepackage{algpseudocode}%
\usepackage{tikz}
\usetikzlibrary{shapes.geometric, arrows, patterns.meta, positioning,calc}

\tikzstyle{plain} = [rectangle,
minimum width=0.5cm,
minimum height=0.5cm,
text centered]

\tikzstyle{arrow} = [thick,->,>=stealth]

\tikzstyle{data} = [rectangle,
minimum width=0.5cm,
minimum height=0.5cm,
text centered,
draw=black]

\tikzstyle{loss} = [rectangle,
minimum width=0.75cm,
minimum height=0.75cm,
text centered,
draw=black]

\tikzstyle{pred} = [rectangle,
minimum width=0.5cm,
minimum height=0.5cm,
text centered,
draw=black]

\tikzstyle{agg} = [circle,
minimum width=0.5cm,
minimum height=0.5cm,
text centered,
draw=black,
thick]

\usepackage{hyperref} 
\usepackage{orcidlink} 
\usepackage[inline]{enumitem} 
\usepackage[export]{adjustbox} 
\usepackage[section]{placeins} 
\usepackage{natbib}

\usepackage{mathrsfs}
\usepackage{xspace}

\newcommand*{\umichaddress}{\orgname{University of Michigan}, \orgaddress{\street{500 South State Street}, \city{Ann Arbor}, \postcode{48109-2125}, \state{Michigan}, \country{USA}}}

\renewcommand{\it}[1]{\textit{#1}}
\newcommand{\tw}[1]{\texttt{#1}}

\newcommand{\sicounter}{
  \renewcommand{\thesection}{S\arabic{section}} %
  \renewcommand{\thefigure}{S\arabic{figure}} %
  \setcounter{figure}{0} %
  \renewcommand{\thetable}{S\arabic{table}} %
  \setcounter{table}{0} %
  \renewcommand{\theequation}{S\arabic{equation}} %
  \setcounter{equation}{0} %
  \renewcommand{\thealgorithm}{S\arabic{algorithm}} %
  \setcounter{algorithm}{0} %
}

\usepackage{amsmath}
\usepackage{amsfonts}
\newcommand{\Rs}{\mathbb{R}}

\newcommand{\set}[1]{\left\{#1\right\}}

\newcommand{\numberthis}{\addtocounter{equation}{1}\tag{\theequation}}
\newcommand{\mc}[1]{\mathcal{#1}}

\newcommand{\X}{\mc{X}}
\newcommand{\Y}{\mc{Y}}
\newcommand{\T}{\mc{T}}
\newcommand{\condset}[2]{ \left\{ #1 \;\middle|\; #2 \right\} }
\newcommand{\zero}{\vec{0}}

\newcommand{\vtheta}{\boldsymbol{\theta}}

\newcommand{\vtau}{\boldsymbol{\tau}}

\DeclareMathOperator*{\argmax}{argmax}
\DeclareMathOperator*{\argmin}{argmin}

\newcommand{\ip}[2]{\langle #1,  #2 \rangle}
\renewcommand*{\vec}[1]{\mathbf{#1}}
\newcommand{\y}{\mathbf{y}}
\newcommand{\m}{\mathbf{m}}

\newcommand{\D}{\mathcal{D}}
\newcommand{\Z}{\mathcal{Z}}
\newcommand{\x}{\vec{x}}

\newcommand{\ve}{\vec{e}}

\newcommand{\eq}[1]{Equation~\ref{eq:#1}}
\renewcommand{\sec}[1]{\nameref{sec:#1}}
\newcommand{\sieq}[1]{Supplementary Equation.~\ref{sieq:#1}}
\newcommand{\sisec}[1]{Supplementary Section~\ref{sisec:#1}}
\newcommand{\fig}[1]{Fig.~\ref{fig:#1}}
\newcommand{\sifig}[1]{Supplementary Fig.~\ref{sifig:#1}}
\newcommand{\sialg}[1]{Supplementary Algorithm~\ref{sialg:#1}}
\newcommand{\tab}[1]{Table~\ref{tab:#1}}
\newcommand{\sitab}[1]{Supplementary Table~\ref{sitab:#1}}

\newcommand{\property}{\tw{property}\xspace}
\newcommand{\scale}{\tw{scale}\xspace}
\newcommand{\all}{\tw{all}\xspace}

\newcommand{\sklearn}{SciKit-Learn\xspace}
\newcommand{\rdkit}{RDKit\xspace}
\newcommand{\pubvinas}{PubViNaS\xspace}

\newcommand{\gendesc}{{BoUTS}\xspace}
\newcommand{\longgendesc}{{\textbf{Bo}osted \textbf{U}niversal and \textbf{T}ask-specific} \textbf{S}election\xspace}
\newcommand{\mutar}{Dirty LASSO\xspace}
\newcommand{\multiboost}{MultiBoost\xspace}
\newcommand{\graphsol}{GraphSol\xspace}
\newcommand{\opera}{OPERA\xspace}

\newcommand{\lgbm}{LightGBM\xspace}

\newcommand{\padel}{PaDEL\xspace}

\newcommand*{\enuma}{\textbf{a)}\xspace}
\newcommand*{\enumb}{\textbf{b)}\xspace}
\newcommand*{\enumc}{\textbf{c)}\xspace}
\newcommand*{\enumd}{\textbf{d)}\xspace}
\newcommand*{\enume}{\textbf{e)}\xspace}

\newcommand{\logp}{logP\xspace}
\newcommand{\logh}{$H_s$\xspace}
\newcommand{\melt}{$T_m$\xspace}
\newcommand{\boil}{$T_b$\xspace}

\newcommand*{\smiles}{SMILES\xspace}
\newcommand*{\nan}{\tw{NaN}\xspace}

\newcommand{\np}{NP\xspace}
\newcommand{\nps}{NPs\xspace}

\newcommand*{\ie}{\emph{i.e.},\xspace}
\newcommand*{\eg}{\emph{e.g.},\xspace}
\newcommand*{\etal}{\emph{et al.}\xspace}

\begin{document}
\newcommand{\finalshorttitle}{Universal Feature Selection}
\newcommand{\finaltitle}{Universal Feature Selection for Simultaneous Interpretability of Multitask Datasets}

\title[\finalshorttitle]{\finaltitle}


\author[1]{\fnm{Matt} \sur{Raymond} \orcidlink{0009-0004-6381-4068}}\email{mattrmd@umich.edu}
\author[2]{\fnm{Jacob Charles} \sur{Saldinger}\orcidlink{0000-0001-6824-8692}}\email{jsald@umich.edu}
\equalcont{Contributed as a member of Chemical Engineering at the University of Michigan}
\author[3]{\fnm{Paolo} \sur{Elvati}\orcidlink{0000-0002-6882-6023}}\email{elvati@umich.edu}
\author[1,4]{\fnm{Clayton} \sur{Scott}\orcidlink{0000-0002-0373-817X}}\email{clayscot@umich.edu}
\author*[1,3,5]{\fnm{Angela} \sur{Violi}\orcidlink{0000-0001-9517-668X}}\email{avioli@umich.edu}

\affil[1]{\orgdiv{Electrical Engineering and Computer Science}, \umichaddress}
\affil[2]{\orgdiv{Low Carbon Pathway Innovation at BP}}
\affil[3]{\orgdiv{Mechanical Engineering}, \umichaddress}
\affil[4]{\orgdiv{Statistics}, \umichaddress}
\affil[5]{\orgdiv{Chemical Engineering}, \umichaddress}


\abstract{
  Extracting meaningful features from complex, high-dimensional datasets across scientific domains remains challenging.
  Current methods often struggle with scalability, limiting their applicability to large datasets, or make restrictive assumptions about feature-property relationships, hindering their ability to capture complex interactions.
  \gendesc's general and scalable feature selection algorithm surpasses these limitations to identify both universal features relevant to all datasets and task-specific features predictive for specific subsets.
  Evaluated on seven diverse chemical regression datasets, \gendesc achieves state-of-the-art feature sparsity while maintaining prediction accuracy comparable to specialized methods.
  Notably, \gendesc's universal features enable domain-specific knowledge transfer between datasets, and suggest deep connections in seemingly-disparate chemical datasets.
  We expect these results to have important repercussions in manually-guided inverse problems.
  Beyond its current application, \gendesc holds immense potential for elucidating data-poor systems by leveraging information from similar data-rich systems.
  \gendesc represents a significant leap in cross-domain feature selection, potentially leading to advancements in various scientific fields.
}


\maketitle

Multitask learning (MTL) is a rich subfield within machine learning that exploits commonalities across tasks (\ie datasets) to build robust and generalizable models.
MTL has been applied to various domains such as natural language processing and computer vision.
MTL trains models on multiple related learning tasks while applying a common regularization to all models (\eg weight sharing).
The underlying idea is that models can share information and representations, leading to better performance on all tasks compared to training the models independently.

MTL plays an important role across various research domains due to the prevalence of datasets that are either large and generic or small and specific.
In fields such as biomedicine, drug discovery, and personalized medicine, predicting individual responses is challenging, as large datasets for common diseases poorly capture specific mutations or rare conditions~\citep{dereli2019, yuan2016, valmarska2017}.

Within this context, multitask feature selection focuses on choosing the most relevant features for these multiple tasks.
Primarily, this approach enhances interpretability, which can help us gain valuable insights into the underlying relationships between different phenomena.
Additionally, selecting features in a structured way improves the generalizability and efficiency of the multitask learning model.
At the same time, it can also improve model performance and reduce computational costs during training and inference.

In this study, we investigate the existence and selection of universal features that are predictive across all datasets under consideration.
The goal is to enable knowledge transfer from well-established research areas to those that are less investigated.
While numerous methods have been devised to select features in multitask problems, they all have limitations that make them poorly suited for this task.

As discussed later, current methods such as Group LASSO ``relax'' universal features to ``common'' features -- those of importance to a subset of tasks -- thereby leading to larger, less interpretable feature sets.
Moreover, tree-based methods~\citep{ke2017, chapelle2011} require a positive correlation among model outputs across tasks, an assumption which may not always hold true.
Additionally, kernel methods~\citep{jebara2004, argyriou2008, peng2016, li2014, dereli2019} face challenges with scalability.
Collectively, such limitations make these approaches ill-suited for application to many real-world datasets.

Therefore, we introduce \longgendesc (\gendesc), a scalable algorithm designed to perform non-linear universal feature selection without restrictions on task structure.
\gendesc identifies universal (common to all tasks) and task-specific features in task subsets, providing insights into unique mechanisms relevant to specific outcomes.

\gendesc has two stages.
First, it performs boosting using ``multitask trees,'' which select universal features based on the minimum feature importance (impurity decrease) across all outputs.
This approach ensures that universal features are predictive for all outputs.
Second, task-specific features are selected by independently correcting each output of the multitask tree using regular boosted trees~\citep{breiman1996b}.
\gendesc penalizes adding new features during tree growth to ensure small feature sets~\citep{xu2014}.
Final predictions are made by summing the individual predictions from the multitask and single-task boosting stages.

We evaluated the performance of \gendesc using seven chemistry datasets that span three distinct molecular classes and employ six different molecular properties for outputs.
\gendesc outperforms existing multitask selection methods in model flexibility, feature sparsity, stability~\citep{nogueira2018}, and an enhanced capacity for selecting universal features.
\gendesc' universal features, even when generalizing across different properties and molecule types, remain competitive and remarkably sometimes surpass specific dataset-tailored methodologies.
Moreover, the identified universal and task-specific features are consistent with established chemical knowledge, highlighting the potential of \gendesc to enhance the analysis of complex datasets and promote the transfer of domain knowledge.

\section*{Results}\label{sec:results}
\subsection*{Overview of \gendesc}\label{sec:overview}

The \gendesc algorithm combines new and existing methods to select concise sets of universal and task-specific features without universal feature relaxation or positive task correlation.
Our two-part strategy is first to select universal features and then select task-specific features (\fig{fig1}a).
This approach is a greedy approximation of a globally optimal feature set, as shown in \sisec{boosting}.
In principle, universal features may be selected using standard Gradient Boosted Trees (GBTs) by independently fitting GBTs on each task and comparing the feature-wise information gained between tasks.
However, feature correlation may cause the independent GBTs to
\begin{enumerate*}[label=\textbf{\arabic*)}]
  \item miss universal features
  \item pick excessive, redundant features.
\end{enumerate*}
To address the first issue, we select universal features using \emph{multitask trees}.
Our multitask trees greedily select features that maximize the minimum impurity decrease across all tasks, ensuring that all trees agree on which (possibly correlated) features to select (\fig{fig1}b).
The second problem is addressed using penalized impurity functions as defined by Xu \etal~\citep{xu2014}.
We penalize the use of new features at each split when selecting universal or task-specific features.
Notably, this approach makes no assumptions about the correlation between task outputs, meaning that \gendesc applies to a wider range of multitask problems than competing methods.
\sialg{boosting} contains the algorithmic details of our approach.
\gendesc's greedy application of multi- and single-task trees and a penalized impurity function underlies its ability to select universal features from nonlinear and uncorrelated tasks.

To quantify \gendesc's performance on real-world data, we evaluate it on datasets covering various chemical scales and properties used to screen molecules for industrial or medical applications.
Our data include seven datasets (\tab{specialized}): four related to small molecules, two to nanoparticles (\np), and one to proteins.
Overall, we consider six different properties: the octanol/water partition coefficient (\logp), Henry's law constant (\logh), the melting (\melt) and boiling (\boil) temperature, solubility in water, and the zeta potential.

\subsection*{Feature selection using \gendesc}
\begin{figure*}
  \centering
  \input{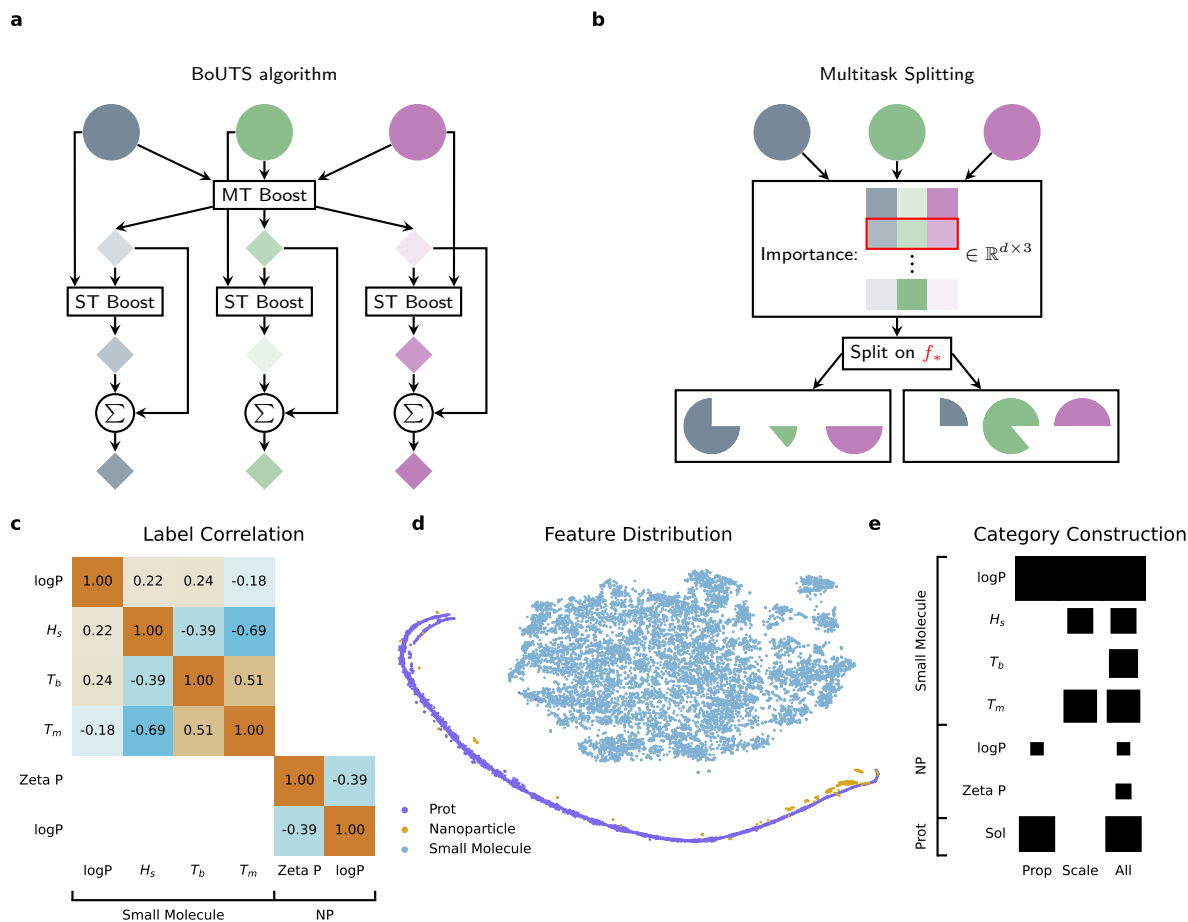}
  \caption{
  \textbf{Overview of \gendesc algorithm and datasets:}
  \enuma illustrates the \gendesc algorithm for the case where $T=3$.
  The boosted multitask trees are trained on all multitask datasets (circles) to estimate (upper diamonds) each task output.
  Single-task boosting estimates the residuals of the multitask trees (middle diamonds).
  We sum over multi- and single-task outputs for the final estimate (lower diamonds).
  In \enumb, we show the splitting process for multitask trees.
  The improvement (impurity decrease) is computed for each task/feature combination (square), and $f_*$ is selected as the feature with the maximin improvement.
  $f_*$ splits each dataset (partial circles), and we repeat until a stopping condition is reached.
  \enumc shows the correlation between dataset outputs.
  Proteins are not included because we use only one protein dataset.
  $n$ values for the lower triangles, grouped by column, are \logp: $n = [777 \; 1{,}185 \; 2{,}143]$, \logh: $n = [479 \; 614]$, \boil: $n = [822]$, zeta potential: $n = [119]$.
  \enumd t-SNE plot of each datapoint (using the complete feature set), colored by molecule type.
  For small molecule, \np, and protein, $n = [11{,}079 \; 3{,}071 \; 234]$, respectively.
  \enume illustrates the assignment of datasets to categories, with square size indicating the logarithm of the size.
  For each dataset (starting at the top row), we have $n = [11{,}079 \; 777 \; 1{,}185 \; 2{,}143 \; 147 \; 206 \; 3{,}071]$.
  }
  \label{fig:fig1}
\end{figure*}

To verify that \gendesc's results are not an artifact of highly correlated tasks, we computed the Spearman correlation of the target properties, as shown in \fig{fig1}c.
The mean absolute correlation of our datasets is low (0.37)~\citep{kaptein2022}, and only two pairs of datasets have an absolute correlation greater than 0.50 (small molecule \melt and small molecule \logh or \boil).
This low positive and negative correlation mixture ensures sufficient differences to render \multiboost ineffective and demonstrate the generality of \gendesc.

Moreover, we visualize the feature-space diversity of our datasets to show that \gendesc's apparent generality is not an artifact of datasets with significant feature-space overlap.
\fig{fig1}d shows a t-SNE plot~\citep{maaten2008} of the union of all datasets computed using the candidate features.
Different types of molecules span distinct regions of the feature space, with small molecules separated from \nps and proteins.
While \nps and proteins partially overlap, the \nps form distinct clumps along the space that proteins cover since our larger protein dataset allows for a more thorough covering of the feature space.
The disjointness of our datasets in feature space suggests that our later results represent the true generalization capabilities of our model.

While \gendesc selects universal features by construction (as they are used only if they improve the performance of every task), there is no guarantee about the number or predictive performance of the selected features.
Therefore, we tested \gendesc's universal features selection on three different ``categories'' of data, \property, \scale, and \all (see \sec{construction} and \fig{fig1}e): the first two categories contain datasets with similar properties or scales, while the third category contains all datasets.

For all the categories, \gendesc selects approximately ten features out of the 1,205 to 1,691 total features (depending on the category, see \sec{construction}).
These universal features are highly predictive, as seen in Figs. \ref{fig:fig2}a-c, despite the large reduction in the number of features (less than 1\% for any category).
Figs. \ref{fig:fig2}a-c show that \gendesc's selected features achieve a similar loss to the original feature set and specialized methods.
The specialized techniques (\tab{specialized}) each use a different, optimized feature space (which may not overlap with ours) and show the performance achievable without performing universal feature selection.
Additionally, we compare the performance of our original feature set, universal and task-specific features, and only universal features on all three categories.
When using universal and task-specific features, the median error is within the interquartile range of specialized methods, even outperforming them on datasets such as \np \logp (Figs. \ref{fig:fig2}a-c).
Compared to specialized methods, \gendesc's feature selection improves cross-dataset interpretability while maintaining a high level of performance.

\gendesc's multitask trees also significantly improve selection stability~\citep{nogueira2018} over simpler greedy approximations of the $\ell_1$ penalty (\fig{fig2}d).
For example, we find no universal features if we run single-task GBT feature selection on each task independently.
Additionally, if we run 100 randomized \gendesc replicates on all seven datasets, the universal features have a stability of 0.36 (higher is better, bounded by $[0,1]$).
By contrast, using GBT feature selection for nanoparticle zeta potential alone achieves a stability of only 0.14.
We find Cohen's $d$ of 18 and a $p$-value of 0.0 (to numerical precision) when using a two-sided Z-test.
Indeed, \gendesc's universal feature selection stability is comparable to that of the Protein dataset, which has 10 times as many samples.
Full tables are available in Supplemental Tables \ref{sitab:universal_summary}, \ref{sitab:task_summary}, \ref{sitab:significance}.
These results demonstrate that \gendesc's multitask trees improve the worst-case feature selection stability compared to similar greedy optimization methods.

Notably, \gendesc's universal features form correlated clusters, suggesting further stability in the mechanisms selected.
We create a correlation matrix for \gendesc's universal features, where each entry indicates the average Spearman correlation between two selected features across datasets.
We ignore datasets where at least one feature in a pair is undefined.
Spectral clustering is performed on this matrix to find five groups of features (\fig{fig2}e).
Three-fifths of the clusters contain at least one feature from each category.
The \all category is represented in all clusters, and the \property and \scale categories are each missing from one cluster.
This overlap indicates that similar information is captured for all categories, even when the selected features don't match exactly.
Thus, we anticipate \gendesc to be even more stable when considering the mechanisms governing the selected features rather than the features themselves.

\subsection*{Comparing \gendesc to other selection methods}\label{sec:compare}

\gendesc is not the first method to model both shared and independent task structures.
However, current approaches are unsuitable for selecting universal features from real-world scientific datasets.
We compare against \multiboost and \mutar as they embody the major shortcomings in the literature.

\multiboost~\citep{chapelle2011} adapts gradient-boosting to model correlated and independent task structures.
The features used to model the correlated task structure may be considered universal.
However, \multiboost cannot model tasks with uncorrelated outputs, so it cannot select universal features for common scientific datasets like ours (\fig{fig1}c).
Indeed, \multiboost models no correlated structure for our datasets and selects only two or three universal features via coincidental overlap.
Additionally, the lack of a sparsity penalty results in the selection of multiple redundant features.
Smaller sets are easier to hold in memory~\citep{miller1956}, so large, redundant feature sets are less interpretable.

\mutar~\citep{jalali2010} employs a ``group-sparse plus sparse''~\citep{rao2013} penalty to linearly model a common (not universal) and task-specific structure.
This approach results in larger feature sets because common features may not describe all tasks and must be supplemented with additional features.
Furthermore, \mutar's linear models may select multiple features that are related by nonlinear function, which increases the feature set even more.

By comparing the feature sets selected by each method, we find that \gendesc selects fewer features than either \multiboost or \mutar while achieving comparable performance.
We perform the same evaluations as in Figs. \ref{fig:fig2}a-c, this time comparing \gendesc and the competing methods in Figs. \ref{fig:fig3}a-c.
To establish an objective comparison, we chose feature penalties that lead to similar performance metrics for each method (see \sec{training} for details).
However, Figs. \ref{fig:fig3}a-c show that \gendesc's feature set is often much smaller than that of either competing method and frequently contains fewer total features than the common features selected by \mutar.
Overall, \gendesc selects fewer total features than its competitors in 10 out of 13 cases, sometimes by a large margin.
For example, despite comparable performance, for \np zeta potential (\fig{fig3}a) \gendesc selects only nine universal features to \mutar's 182 and \multiboost's 72.

Figs.~\ref{fig:fig3}a-c shows that competing methods only outperform \gendesc when they select significantly more features, like for \np \logp, where the number of features differs by nearly two orders of magnitude.
We presume that this difference is caused by \multiboost's lack of an explicit sparsity term and \mutar's linearity assumption.
Nevertheless, this combination of similar performance and smaller feature sets suggests that \gendesc is more capable of providing insight into the physical mechanisms controlling the feature-property relationships of interest.
This comparison shows that \gendesc is significantly more interpretable than competing methods.

Additionally, we observe that the number of task-specific features selected by \gendesc closely aligns with theoretical assumptions regarding universal feature behavior.
\gendesc generally selects fewer task-specific features when a task has fewer samples.
As described in \sec{universal_splitting}, \gendesc adds features to the universal set by repeatedly splitting on the feature that maximizes the minimum information gain (prediction improvement) across all tasks.
Adding new features is penalized, so \gendesc selects universal features only if they are predictive for \it{all} tasks.
Because weak feature-property relationships may only be detectable in larger datasets, the smallest datasets will limit the universal feature set.
However, tasks with larger datasets will compensate by selecting additional task-specific features, which is the desired behavior of universal and task-specific feature selection.

\subsection*{Analysis of selected features}
Analysis of \gendesc's task-specific features showcases its ability to recover well-known and quantitative feature-property relationships.
For example, selecting the ``total molecular mass'' feature for small molecule \boil reflects an established correlation connected to intermolecular dispersion forces~\cite{wessel1995}.
Similarly, \gendesc selects solvent-accessible surface area for small molecule \logp, commonly used to compute non-polar contributions to the free energy of solvation~\cite{liu2019}.
Such findings illustrate that \gendesc's sparse feature selection can recover scientifically meaningful feature-property relationships.

Further, we find that \gendesc's features are highly specialized to the types of molecules and properties under consideration.
We note that half of the universal features for the \scale category are VSA style descriptors~\cite{labute2000}, which are not universal for either the \property or \all categories.
This observation corroborates previous findings on VSA descriptors' effectiveness for solubility and \boil in small molecules~\cite{labute2000}.
Additionally, their absence from the \property and \all categories is expected since VSA descriptors are not anticipated to apply to other length scales or molecular motifs.

Finally, we find that \gendesc's universal features provide insight into factors controlling chemical properties across multiple scales.
When analyzing the \property category, we note the lack of task-specific features for \np, which suggests that the universal features are sufficiently descriptive for \logp predictions of metal-cored \nps.
Although an \np's solubility may partially depend on its metal core, metal-related features are weakly predictive for small molecules and proteins.
Similarly, tessellation descriptors~\cite{yan2020}, linked to carbon, nitrogen, and oxygen atom subgroups, are the only extrinsic descriptors chosen for the \property category.
If we compute the information gain provided by each feature the ensemble uses, we find that tesselation features contribute 46\% of the total information gain.
Because the model's competitive \np \logp predictions heavily utilize tessellation, we suspect the model focuses on the \nps' organic ligand surface, which is in contact with the solvent.
A researcher may correctly conclude from these results that ligating known soluble molecules to a nanoparticle will improve the nanoparticle's solubility.

\section*{Discussion}\label{sec:discussion}

\gendesc breaks new ground in cross-domain feature selection, overcoming the limitations of existing methods that often struggle with the intricate complexities of real-world data, particularly nonlinearities and uncorrelated tasks.
Through our case study, we find that \gendesc selects far fewer features than existing methods (typically less than $1\%$ of features) while achieving similar performance.
We find that approximately ten universal features are predictive for all tasks, and that task-specific features further improve performance (Figs. \ref{fig:fig2}a-c).
\gendesc exhibits desirable sparsity patterns, such as selecting more task-specific features for tasks with larger datasets.
This remarkable sparsity enhances interpretability, especially in smaller datasets where understanding the underlying drivers is crucial.
By focusing on the essential features, \gendesc empowers researchers to gain deeper insights into their data, fostering a more comprehensive understanding of the scientific phenomena at play.

We attribute \gendesc's performance to its universal, nonlinear, and more general approach to feature selection.
LASSO-like methods, such as \mutar, relax universal features to ``common'' features, so more features must be selected to compensate for underrepresented tasks.
Additionally, linearity likely leads to the redundant selection of non-linearly-related features.
Similarly, current tree ensemble methods like \multiboost require datasets to follow restrictive assumptions, such as having correlated outputs.
Since our dataset outputs are not correlated, \multiboost cannot be applied.
\gendesc performs nonlinear feature selection with neither relaxation nor assumption, leading to sparser feature sets without performance degradation.

As demonstrated by our results, \gendesc's unique ability to select universal features from uncorrelated tasks holds immense potential for uncovering unifying principles across diverse scientific domains.
For example, \gendesc correctly identifies the connection between surface chemistry and solubility in proteins, nanoparticles, and small molecules without any \emph{a priori} knowledge of chemistry.
Beyond solubility, our universal features are predictive across all seven datasets.
Such results suggest that similar \gendesc-derived connections can be made in many real-world problems.

Although mayn works discuss chemistry datasets and how they relate to the underlying ``chemical space,'' our universal feature selection is most comparable to Medina-Franco \etal's ``consensus chemical space''~\citep{medina2022}.
Given multiple types of molecules with different chemical features, this space is composed of pooled versions of the original features.
The thesis is that a single set of descriptors is ``not enough'' for many prediction tasks.
Our universal features contradict this assertion, as they are important for \emph{all} datasets under consideration.
Since our universal feature set is performant, small, and uses standard chemical features, we expect that they will provide a more interpretable and computationally-efficient alternative to the consensus chemical space.
Indeed, the universal feature selection suffer a relevant predictive penalty, as high-performing universal features exist for all three categories of chemistry datasets we evaluated.

\gendesc's universal features allow domain-specific knowledge transfer between datasets.
Notably, applying universal feature selection to small datasets improves the stability of the selection ($p=0$ to numerical precision).
This leads to greater confidence that the selected features are optimal for the true data distribution~\citep{lee2013}.
Although the universal features differ between categories, they encode similar information (\fig{fig2}d).
Since we require universal features to be important for \emph{all} tasks, one might anticipate that using more datasets during selection would reduce the number of clusters represented by the universal features.
Thus, the ubiquity of the \all category (\fig{fig2}d) along with the similar performance of core features across categories (Figs. \ref{fig:fig2}a-c) suggests a deeper connection between chemical datasets.
Such a connection raises an interesting question: how well do universal features generalize to new datasets.

The answer is strongly dependent on the field and data under consideration, but when universal features do generalize, there are several immediate benefits.
Consider settings where gathering features is expensive, such as assays, gene sequencing, or data surveying.
Here, researchers may save time and money by only collecting universal features that have been selected using similar, existing datasets.
Moreover, generalizable universal features may have exciting connections to ``foundation models,'' which learn general representations from many related datasets.
Universal features may provide an interpretable alternative that elucidates hidden relationships between datasets (\eg proteins and nanoparticles~\cite{saldinger2023}).

Universal feature selection opens the door to several intriguing possibilities.
In fields where large and small datasets are common, universal feature selection can improve selection stability on small datasets.
Intriguingly, universal features could prove even more useful for inverse problems: universal features may be used to anticipate which modifications will unintentionally affect other properties.
Such foresight could reduce the time, money, and human costs necessary in product research, like drug design.
Similarly, one might select features that are important for one task but uninformative for all others.
These ``unique features'' could be used to identify which structural modifications would have little to no effect on the other properties being considered.

While \gendesc offers significant advantages, it is crucial to acknowledge its limitations and ongoing research efforts.
\gendesc's primary limitation is its reliance on greedy optimization.
Although it outperforms existing methods, \gendesc's greedy feature selection, greedy tree growth, and boosting may still result in a feature set that is larger or less predictive than the optimal feature set.
Additionally, our multitask trees require that each task is split using the same features in the same nodes, which is a seemingly restrictive assumption.
However, boosting reduces the bias of the base learner~\citep{bartlett1998}, so further evaluation is required to determine whether the aforementioned restriction actually harms performance.
Additionally, \gendesc is more computationally expensive than linear methods like \mutar, which may be significant for large datasets.
However, \gendesc is still significantly more scalable than kernel methods.

While demonstrating remarkable performance in our chemical datasets, the true potential of \gendesc lies in its scalability and broader applicability.
Ongoing research efforts will optimize \gendesc using histogram trees, GPU acceleration, and multiprocessing~\citep{chen2016, ke2017}, paving the way for massive datasets with millions of features and samples, a hallmark of modern scientific research.
This scalability will enable the analysis of previously intractable datasets, unlocking a wealth of hidden knowledge across diverse fields, including biology, materials science, finance, and beyond.
By leveraging \gendesc's scalability, researchers can gain deeper insights, discover hidden patterns, and accelerate scientific progress across various fields.

In summary, \gendesc offers unique sparsity, interpretability, and universal feature selection advantages.
Its universal features facilitate the transfer of domain knowledge between datasets and improve feature selection stability on small datasets.
Moreover, \gendesc's universal features perform competitively despite being predictive for all datasets.
We anticipate these capabilities will empower researchers to analyze high-dimensional or poorly understood datasets, accelerating progress across various scientific fields.

\section*{Methods}\label{sec:methods}

\gendesc performs universal and task-specific feature selection in two stages;
first, the universal features are selected using multitask trees, and then the task-specific features are individually selected for each task using single-task trees.
In the second stage, the single-task trees approximate the residuals of the first stage.
The outputs of all trees for each task are summed to make a prediction.
Here we provide the splitting criteria used for single-task and multitask trees.
Full details of the boosting process are provided in \sisec{boosting}.

\subsection*{\gendesc}

\subsubsection*{Splitting single-task trees}
\label{sec:task_splitting}

For selecting task-specific features, we use Xu \etal's~\cite{xu2014} adaptation of the classic CART splitting criteria from \citep{breiman1996b}.
Because the universal feature selection extends this algorithm, we reproduce the details here as a precursor to \sec{universal_splitting}.

The CART algorithm defines the notion of impurity, which is used to grow either classification or regression trees.
Only regression trees are necessary for gradient boosting as we learn the real-valued loss gradients.
Intra-node variance is the canonical example of regression impurity~\citep{breiman1996b}, but one can use any convex function that achieves its minimum when the predicted and ground-truth values match exactly.
From notational simplicity, we use $i(\eta)$ to denote the impurity evaluated on node $\eta$ from this point forward.
This impurity function can then be used to perform greedy tree growth.

The CART algorithm grows trees by recursively performing splits that greedily reduce the impurity of each node.
Let $\eta_{f > v},\eta_{f \leq v}$ represent the right and left child nodes split on feature $f \in F$ at location $v \in \Rs$ for a $d$-dimensional feature set $F$.
Further, let $|\cdot|$ indicate the cardinality of a set and $2^S$ indicate the power set of $S$.
Then, the \emph{impurity decrease}~\citep{breiman1996b} of that split is defined as
\begin{equation}
  \Delta i(\eta, f, v) \doteq i(\eta) - \frac{|\eta_{f \leq v}|}{|\eta|}i(\eta_{f \leq v}) - \frac{|\eta_{f > v}|}{|\eta|}i(\eta_{f > v}) \; ,
\end{equation}
where $\Delta i: 2^{\D_t} \times F \times \Rs \to [0, \infty)$
and the optimal split is the tuple $(f,v)$ that provides the largest impurity decrease.
We start from the root node, which contains all samples and has maximum impurity.
Then, the tree is grown by iteratively applying the optimal split until a stopping criterion is reached.
In practice, splitting is usually stopped once $\Delta i(\eta, f, v)$ or $|\eta|$ fall below a predefined threshold or once the tree reaches a predefined depth or number of leaf nodes.

We add a greedy approximation of the $\ell_1$ penalty to the impurity term to minimize the number of features utilized.
CART trees may be used to select features, but they select many redundant features since there is no feature selection penalty.
Thus, \cite{xu2014} modifies the CART algorithm to use a \emph{penalized impurity function}.
Let $\mathbbm{1}_f^t$ indicate whether feature $f$ is \emph{not} used by a tree in task $t$.
Then, the penalized impurity is
\begin{equation}\label{eq:xu}
  i^t_b(\eta) \doteq i(\eta) + \lambda^t\sum_{f \in F} \mathbbm{1}_f^t
\end{equation}
for boosting round $b$ of task $t$ on node $\eta$.
Then, the \emph{penalized impurity decrease} at a split is defined as
\begin{align*}\label{eq:impurity}
  \Delta i^t_b(\eta, f, v) & \doteq \Delta i(\eta, f, v) - \lambda^t\sum_{f \in F} \mathbbm{1}_f^t \numberthis
\end{align*}
with $\Delta i^t_b: 2^{\D_t} \times \set{1,...,d} \times \Rs \to [0, \infty)$.
For a given node $\eta$, we choose the split feature $f$ and location $v$ that result in the maximum information gain (Figs.~\ref{fig:fig1}b, \sialg{tssc}).
Note that a penalized impurity decrease is no longer restricted to $[0, \infty)$.
This is not an issue in practice because of the threshold for minimum impurity decrease.
In practice, we use MSE with an improvement score~\citep{friedman1999}~(Equation 35) for selecting splits.
The Friedman MSE attempts to maximize the difference between node predictions while maintaining an equal number of samples per node, which is known to improve performance in some settings~\citep{pedragosa2011}.
This penalized impurity decrease enables sparse task-specific feature selection using GBTs.

\subsubsection*{Splitting multitask trees}
\label{sec:universal_splitting}

Universal feature selection requires additional modifications to Breiman~\etal's CART trees.
For a feature to be universal, it must simultaneously be selected by all $T$ tasks during greedy tree growth.
Hence, we assume all $T$ trees from a single boosting round are nearly identical.
The only difference is that each tree may choose its splitting location for the feature used in each non-terminal node.
Thus, we must derive a splitting criterion that
\begin{enumerate*}[label=\textbf{\arabic*)}]
  \item chooses a universal feature to split all task trees on,\label{enum:universal}
  \item allows each task tree to choose its splitting threshold for the universal features.\label{enum:threshold}
\end{enumerate*}

We ensure universal features are important to all tasks via maximin optimization.
Let $\eta^t \subseteq \D^t$ indicate the data points in a tree node currently under consideration.
At boosting round $b$ we find the feature $f_*$ and split locations $v^1_*,...,v^T_*$ that solve
\begin{align}\label{eq:maximin}
  \max_{f \in F} \min_{t \in \set{1,...,T}} \max_{v^t \in \Rs} \Delta i^t_b(\eta^t, f, v^t),
\end{align}
where $v^t$ is a split location on feature $f$ for task $t$.
Note that the split location is found independently for each task.
This method selects the feature that maximizes the minimum impurity to decrease across all tasks while allowing unique split locations, covering requirements \ref{enum:universal} and \ref{enum:threshold} from above.
This approach ensures that all universal features are important for all tasks while keeping the selected universal feature set concise.

\subsection*{Datasets}

\subsubsection*{Construction}\label{sec:datasets}
We chose datasets to ensure wide coverage of multiple properties and molecular scales: small molecules, proteins, and larger metal \nps.
Our small molecule \logp dataset was taken from \citep{popova2018}, which provided \smiles and associated experimental logP values.
For small molecule \logh, \boil, and \melt, we used the 2017 EPISuite~\citep{physprop2014} to extract experimentally-measured properties using the \smiles from the \logp dataset.
During pre-processing, we log-transform $H_s$ to normalize the distribution.
Not every \smiles was associated with a value for all properties, so \smiles from the latter three datasets form overlapping subsets of the \logp \smiles.
The cardinalities of these sets are detailed in \fig{fig1}e.
For proteins, we use solubility values from \citep{han2019} and PDB structures from the Protein Data Bank~\citep{berman2003} and AlphaFold Protein Structure Database~\citep{jumper2021}.
To create the \np dataset, we use the \pubvinas~\cite{yan2020} database to obtain \logp and zeta potential properties and structural information as PDB files.
Overall, this resulted in seven different datasets across three different molecular scales.

Similarly, our full feature set was designed to capture as much nanoscale chemistry as possible.
Such features were based on the radius, solvent accessible surface area, Van der Waal's surface area, atomic property distributions~\cite{isayev2017}, depth from the convex hull of the molecule, WHIM descriptors~\cite{todeschini1997}, and tessellation descriptors~\cite{yan2020}.
Atomic weightings for these descriptors were similar to those used in previous works~\cite{isayev2017, yan2020}.
Notably, not all features are computable for all molecular scales.
For example, a feature describing volume will return \nan when run on a flat molecule.
However, we don't want to universally exclude such features, as they may be useful if defined for all datasets in a category.
Instead of dropping such features from all datasets, we individually drop features that are \nan or constant for each dataset.

\subsubsection*{Construction of categories}\label{sec:construction}
We perform ablation tests on \gendesc by evaluating its performance on three \emph{categories} of datasets, as shown in Figs.~\ref{fig:fig2}a-c: \property, \scale, and \all.
\property contains three datasets with similar solubility-related target properties but span different chemical spaces.
\scale contains three datasets that span similar chemical spaces of small molecules but have different target properties.
\all contains all seven datasets covering all chemical spaces and target properties used in this work.
We use well-defined and non-constant features for all datasets as the candidate feature set for each category.
These categories constitute an ablation test for \gendesc's generalization capabilities.

\subsection*{Performing splits with overlapping datasets}\label{sec:split}
In some (but not all) cases, the same molecule occurs in multiple datasets, so we perform a modified stratified split.
We perform stratified splits on the overlaps of each dataset, which allows us to use all available data while preserving the correct train/validation/test split ratios for all datasets and preventing data leakage.
Because the overlap of dataset samples depends on the datasets under consideration, the same dataset may be split differently when included in different categories.
For example, small molecule \logp is partitioned differently in each category since it overlaps with different datasets (see $n$ in Figs.~\ref{fig:fig2}a-c).
To address this issue, we re-evaluate competing methods for each category using the same splits described here.

\subsection*{Statistical comparison of stabilities}
We use \citep{nogueira2018}'s measure of stability and a statistical test to quantify the selection stability of \gendesc.
For \gendesc's universal feature selection and single-task gradient boosted feature selection, we perform feature selection on $M=100$ randomized train/validation/test splits.
We then encode the binary masks for universal features and each of $T$ independent task in matrices $\Z^U,\Z^1,\ldots,\Z^T \in \{0,1\}^{M \times d}$, where $d$ is the number of features.
In this encoding, $\Z_{m,f}$ indicates whether feature $f$ was selected during the $m$th trial for matrix $\Z$.
Define $\hat{p}_f(\Z) \doteq \frac{1}{M} \sum_{i=1}^{M} \Z_{i,f}$ as the empirical probability of selecting feature $f$, and $s_f^2(\Z) \doteq \frac{M}{M-1} \hat{p}_f(\Z) (1 - \hat{p}_f(\Z))$ as ``the unbiased sample variance of the selection of the $f$th feature.''
Then, \citep{nogueira2018} defines an estimate of \emph{feature selection stability} as
\begin{equation}
  \hat{\Phi}(\Z) = 1 - \frac{1}{d} \sum_{f=1}^{d} s_f^2(\Z) \quad.
\end{equation}
The $(1-\alpha)$\% confidence interval is computed as
\begin{equation}
  \hat{\Phi} + F^{-1}_\mathcal{N}\left(1-\frac{\alpha}{2}\right) \sqrt{v(\hat{\Phi})} [-1, 1],
\end{equation}
for the standard normal cumulative distribution function $F_\mathcal{N}$ and closed interval $[-1,1]$.
To compare the stability of the two models, we compute $\Z^U$ and $\Z^t$ as above, then calculate the test statistic for two-sample, two-sided equality of means~\citep{nogueira2018}.
Letting $v(\hat{\Phi}(\Z))$ indicate the variance estimate of $\hat{\Phi(\Z)}$ (defined in \citep{nogueira2018}), we have the test statistic
\begin{equation}
  T_M = \frac{\hat{\Phi}(\Z^U) - \hat{\Phi}(\Z^t)}{\sqrt{v(\hat{\Phi}(\Z^U)) + v(\hat{\Phi}(\Z^t))}} \quad,
\end{equation}
where $T_M$ asymptotically follows a normal distribution.
Thus, the $p$-value easily computed as $p \doteq 2(1-F_\mathcal{N}(|T|))$~\citep{nogueira2018}.

To quantify the effect size between \gendesc and independent gradient boosted feature selection, we compute Cohen's $d$~\citep{cohen1988} for task $t$ as
\begin{equation}
  \frac{\sqrt{2}(\hat{\Phi}(\Z^U) - \hat{\Phi}(\Z^t))}{\sqrt{v(\hat{\Phi}(\Z^U)) + v(\hat{\Phi}(\Z^t))}}
\end{equation}
since we take an equal number of samples for each model.

\subsection*{Implementation of all methods}\label{sec:specialized}
We utilized community implementations of each method when possible and reimplemented or modified algorithms when necessary.
For \mutar~\citep{jalali2010}, we used the community implementation by \citep{janati2019}.
Since the original \mutar implementation~\citep{janati2019} requires every task to have the same number of samples, we modified the code to mask the loss for outputs whose true label/output is undefined (\eg protein \melt), preventing them from affecting the parameter updates.
For \multiboost~\citep{chapelle2011}, we used \sklearn trees to implement it from scratch.
We use the official \graphsol~\citep{chen2021} implementation;
however, we change the output activation from sigmoidal to exponential since our protein solubility dataset is defined over $(0,\infty)$ rather than $(0,1]$. 
We implement the models for \pubvinas~\citep{yan2020} and \opera~\citep{mansouri2018} using \sklearn~\citep{pedragosa2011}.
\ pubvinas provide the \np features, and we compute the \padel~\citep{yap2011} version 2.21 descriptors for \opera because our datasets do not overlap exactly.
We used the standard \rdkit 2022.03.3 implementation of Crippen \logp~\citep{landrum2022}.

\gendesc was implemented by modifying \sklearn's gradient boosted trees.

\subsection*{Training and evaluation}\label{sec:training}

All feature selection methods were trained across different selection penalties to create regularization paths.
For each category, we perform a randomized 0.7/0.2/0.1 train/validation/test split using the method from \sec{split} and rescale the feature vectors and labels such that each training dataset has a mean of 0 and a standard deviation of 1.
Then, we perform a sweep over the penalty parameters on the training set for each selection method.
For \multiboost, we evaluate $2^n$ models for ${n \in \{0,...,8\}}$ to limit the number of features its trees select.
For \gendesc, we set all universal and task-specific feature penalties equal to $\lambda_n$ and tested penalties that were equally spaced in log-space, with regularization $\log \lambda_n \doteq \frac{8(n-1)}{19}-4$ for $n \in \{1,...,20\}$.
\mutar required a more fine-tuned balance between common and task-specific penalties, so we similarly took equally-log-spaced penalties such that $\log \lambda^s_n \in [-4,-1]$ and $\log \lambda^b_n \in [-3.85, -1]$, where $n=20$ and $\lambda^s_n, \lambda^b_n$ indicate the sparse and group-sparse penalties, respectively.
This approach creates a 2d regularization path for each method, which we later use to select a feature selection penalty.

We use a unique approach to extract the selected features from each model in each regularization path.
For \multiboost, we consider a feature to be universal if it is used by at least one tree in each task or the correlated task structure.
\gendesc is similar, but we consider the multitask boosted trees instead of the correlated task structure.
For both methods, a feature is a task-specific feature if used on at least one but not all tasks.
For \mutar, we consider a feature common if the group-sparse weight matrix uses it and task-specific if it is used only in the sparse weight matrix.
We use a weight threshold of $\varepsilon=10^{-4}$ to determine membership.
This provides the features selected by each model in the regularization path.

We then choose a cutoff point to provide similar performance across all feature selection methods.
For every subset of features selected for each model (9 + 20 + 20 subsets total), its performance is measured using \lgbm to learn and predict the training set using only the selected features.
For each model, we chose the feature penalty that directly precedes a 10\% decrease in the explained variance on any task/training dataset.
We then find the optimal inference model by performing a grid search with \lgbm on the training and validation sets, and we report performance based on the test set.
Because the original labels for each dataset exist on different scales, we report the absolute error of the rescaled labels (0 mean and unit variance), which we call the \emph{normalized absolute error}.
This results in similar performance for each selection method and allows us to concentrate only on feature sparsity.

Finally, we evaluate the specialized, application-specific methods on the same datasets to ensure a fair comparison.
To train \opera and \pubvinas, we use a cross-validated grid search on the union of training and validation sets.
Per model selection for \graphsol was not computationally feasible, and Crippen \logp is not a learning-based method.
We find that \graphsol is missing one protein from our training dataset (\tw{cfa}), so this protein is skipped during \graphsol training.
Prediction is performed on the same hold-out tests as the feature selection methods.

\subsection*{Language model}
We used GitHub Copilot as a programming assistant and ChatGPT4 to generate simple functions.
GitHub Copilot was also used to assist in \LaTeX\ typesetting.
ChatGPT4 and Claude2 were used to review the manuscript's early drafts and brainstorm ideas, including the name ``\gendesc.''

\section*{Data availability}\label{sec:data_availability}
Additional data will be available at Deep Blue Data, an open and permanent data repository maintained by the University of Michigan.
During review, the data will be accessible using a public Google Drive folder.
This repository contains all raw files too large to attach to the paper, including computed features and predictions.
The source data for Figs.~\ref{fig:fig1}, \ref{fig:fig2}, \ref{fig:fig3} are available with this paper.

\section*{Code availability}\label{sec:code_availability}
Publicly hosted implementations of all evaluated code are currently hosted on the University of Michigan GitLab server under the GPL3 license.
The repository will be made public at the time of review, and we will release a CodeOcean capsule with an associated DOI before publication.
The GitLab repository includes an issue tracker to provide feedback.

\section*{Acknowledgements}\label{sec:acknowledgements}
This work was supported in part by the BlueSky Initiative, funded by The
University of Michigan College of Engineering (principal investigator, the Army Research
Office MURI (grant no. W911NF-18-1-0240), and ECO-CBET Award Number (FAIN):
2318495.
We acknowledge Advanced Research Computing, a division of Information and Technology Services at the University of Michigan, for computational resources and services provided for the research.

C. Scott was partly supported by the National Science Foundation under award 2008074, and by the Department of Defense, Defense Threat Reduction Agency under award HDTRA1-20-2-0002.

\section*{Author contributions}
A.V., C.S., and P.E. conceived and supervised the project.
J.S. and P.E. conceived chemical features and representations.
M.R. designed, trained, and tested the machine learning models.
C.S. supervised the formalization of \gendesc.
J.S. helped design computational experiments/evaluations and collected the datasets.
M.R. and J.S. chose the competing methods.
All authors read, revised, and approved the manuscript.

\section*{Competing interests}
The authors declare that they have no competing interests.

\section*{Tables}

\begin{table}[hbt]
  \caption{
    \textbf{Specialized Methods:}
    Specialized, application-specific methods for each property and system combination.
    Horizontal lines group the system and property for each specialized method.
    Note that these differ from the competing feature selection methods.
  }\label{tab:specialized}
  \begin{tabular}{@{}rll@{}}
    \toprule
    \textbf{System}                 & \textbf{Property} & \textbf{Specialized Method}                  \\
    \midrule
    \multirow{4}{*}{Small Molecule} & \logp             & Crippen~\citep{wildman1999}                  \\
    \cmidrule{2-3}
                                    & \logh             & \multirow{3}{*}{\opera~\citep{mansouri2018}} \\
                                    & \melt             &                                              \\
                                    & \boil             &                                              \\
    \midrule
    \multirow{2}{*}{Nanoparticle}   & \logp             & \multirow{2}{*}{\pubvinas~\citep{yan2020}}   \\
                                    & Zeta Potential    &                                              \\
    \midrule
    Protein                         & Solubility        & \graphsol~\citep{chen2021}                   \\
    \botrule
  \end{tabular}
\end{table}

\section*{Figures}

\begin{figure*}[t]
  \centering
  \includegraphics[width=\textwidth]{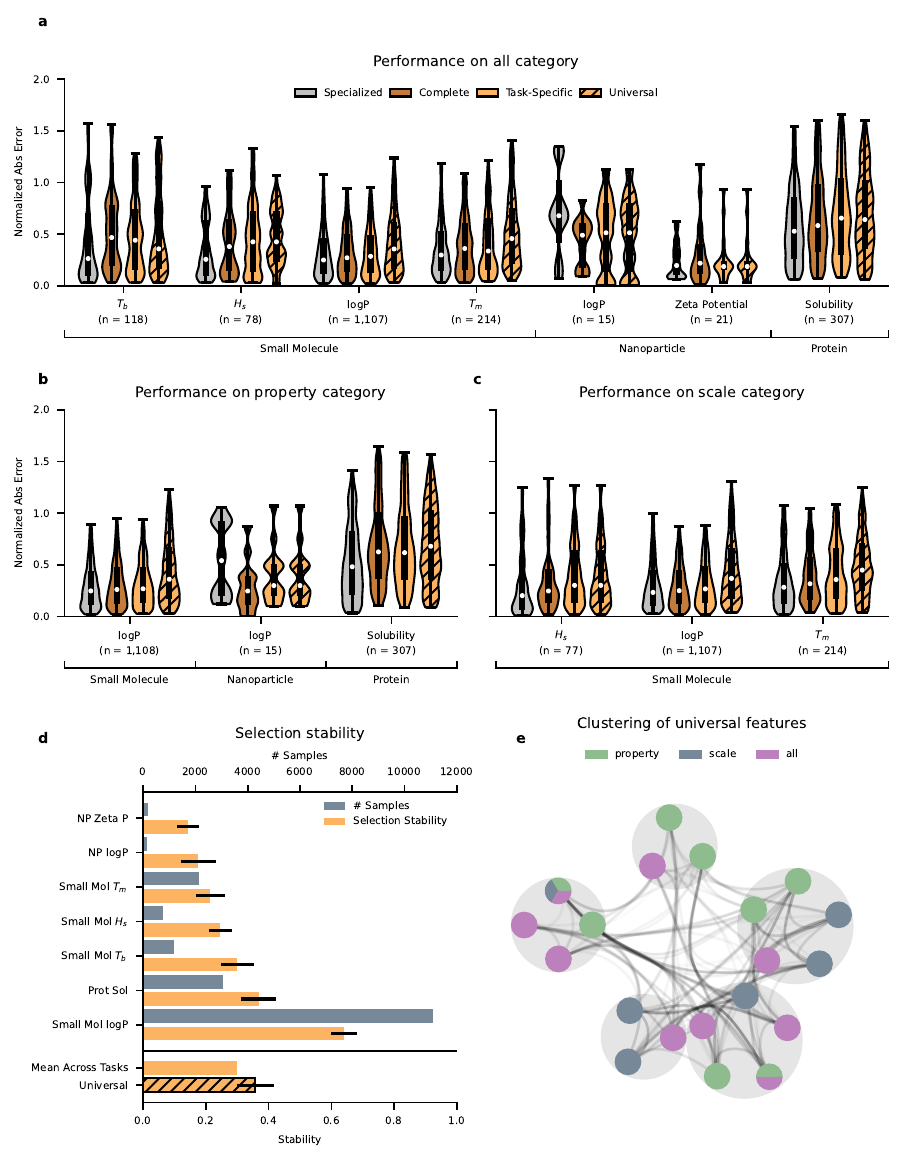}
  \caption{
    \textbf{Ablation tests and analysis of \\gendesc selected features:}
    Feature ablation test for \gendesc are shown in \enuma, \enumb, and \enumc, and compare our selected features to specialized prediction methods.
    Violin plots show the performance distribution; the inner bars indicate the 25th and 75th percentiles, and the outer bars indicate the 5th and 95th percentiles.
    The white dot indicates the median performance.
    The top of \enumd shows the dataset size (top axis) and the selection stability of single-task gradient-boosted feature selection (bottom axis).
    The bars indicate the 95\% confidence interval.
    In the bottom section, the upper bar shows the mean stability across all tasks.
    The lower bar shows the stability of \gendesc's universal features, with the 95\% confidence interval as a black bar.
    \enume shows the absolute Spearman correlation between the universal features as a graph, with clusters indicated by gray circles and node colors indicating the categories that selected that feature.
  }
  \label{fig:fig2}
\end{figure*}

\begin{figure*}[t]
  \centering
  \includegraphics[width=\textwidth]{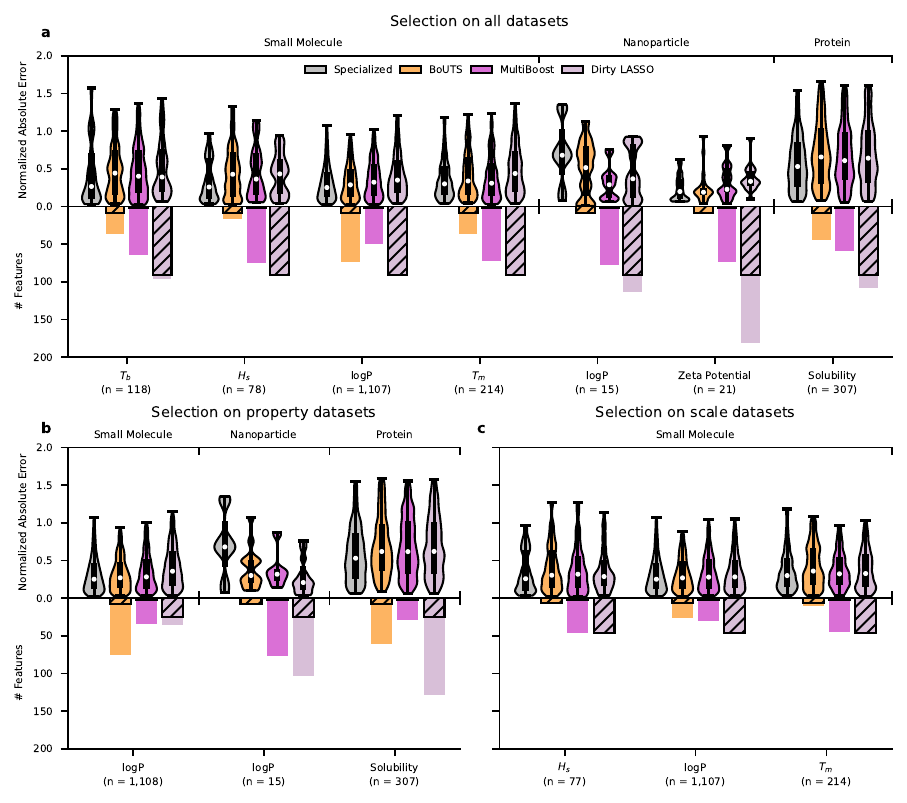}
  \caption{
    \textbf{Comparing the performance of \gendesc and competing selection methods:}
    The top half of \enuma shows the performance of all evaluated feature selection algorithms compared to specialized methods.
    The violin plots are defined in \fig{fig2}a.
    The bottom half shows the number of features selected by each method.
    No features are indicated for the specialized methods as they are not selection methods.
    The hatched section indicates the universal or common features that are selected, and the remaining features are task-specific.
    Plots \enumb and \enumc are defined similarly for the \property and \scale categories.
  }
  \label{fig:fig3}
\end{figure*}


\clearpage

\onecolumn
\begin{appendices}
  \sicounter

\section{\gendesc selected universal features}
\label{sisec:features}

In \sitab{universal_features}, we list the universal features selected by \gendesc for each category.
We do not include the task-specific features, as there are too many, and they are provided with the associated code.

Note that \gendesc's feature selection may not be stable across multiple runs of the same model or different splits of the same datasets.
However, \gendesc enables intertask-comparisons of selected features by selecting universal features that are important for all tasks and task-specific features that are important for at least one task.
Thus, selected features may be unstable across categories, and each category may select different features from a set of correlated features.

\begin{table}[h]
  \centering
  \caption{
    \textbf{Universal features selected for each category:}
    They are presented in alphabetical order and grouped across categories.
  }
  \label{sitab:universal_features}
  \begin{tabular}{lll}
    \toprule
    \textbf{Property}                       & \textbf{Scale}                 & \textbf{All}                             \\
    \midrule
    \tw{atomic\_volume\_cccc}               & \tw{atomic\_volume\_cccc}      & \tw{atomic\_volume\_cccc}                \\
    \tw{atomic\_volume\_cccn}               &                                &                                          \\
                                            &                                & \tw{atomic\_volume\_ccco}                \\
    \tw{atomic\_volume\_ccno}               &                                &                                          \\
    \tw{atomic\_volume\_oooo}               &                                &                                          \\
                                            & \tw{covalent\_radius\_2}       &                                          \\
                                            & \tw{covalent\_radius\_3}       &                                          \\
                                            &                                & \tw{electron\_affinity\_mean}            \\
    \tw{electron\_affinity\_variance}       &                                &                                          \\
                                            &                                & \tw{electron\_affinity\_whim\_axis2}     \\
                                            &                                & \tw{electron\_affinity\_whim\_d}         \\
    \tw{electronegativity\_mean}            &                                &                                          \\
                                            &                                & \tw{electronegativity\_variance}         \\
                                            &                                & \tw{evaporation\_heat\_convhull\_median} \\
                                            & \tw{group\_5}                  &                                          \\
                                            & \tw{ionization\_energy1\_mean} &                                          \\
                                            &                                & \tw{ionization\_energy1\_variance}       \\
                                            & \tw{mass\_ccno}                &                                          \\
    \tw{polarizability\_whim\_ax\_density3} &                                &                                          \\
    \tw{vdw\_volume\_variance}              &                                & \tw{vdw\_volume\_variance}               \\
    \bottomrule
  \end{tabular}
\end{table}

\section{Universal feature stability}
\label{sisec:stability}

\begin{figure}
  \centering
  \includegraphics[width=0.75\textwidth]{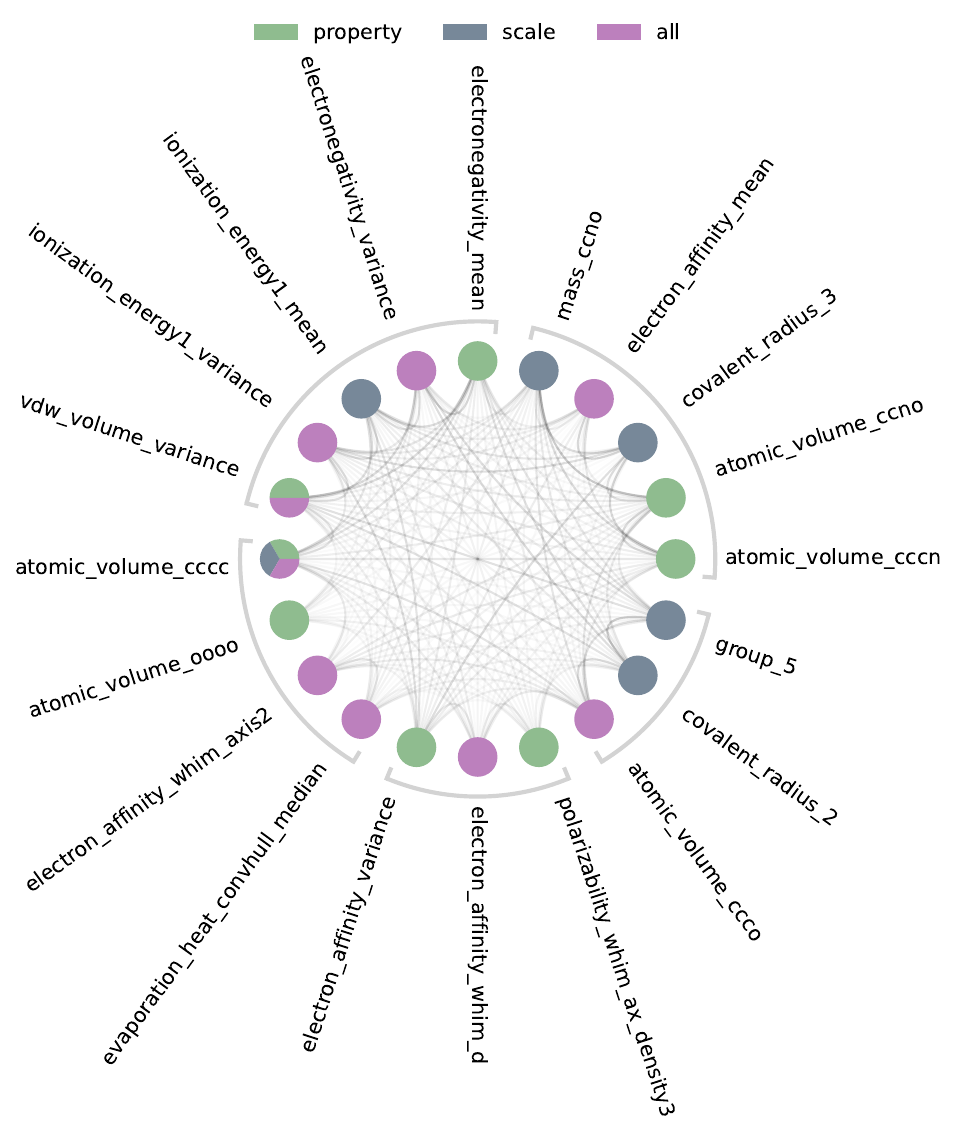}
  \caption{
    Alternative visualization of \fig{fig2}e with included feature names.
    The edges indicate the absolute Spearman correlation between the universal features selected for each category, with clusters indicated by circular brackets on the outside of the graph.
    The colors in each node indicate the categories that selected that feature.
  }
  \label{sifig:fig2_si}
\end{figure}

Here, we provide stability metrics computed for \gendesc and single-task GBT feature selection.
Stability and variance are computed using code from \citep{nogueira2018}.
All hyperparameters are the same for all models used in this table, except for the feature penalty.
The penalty was adjusted to provide approximately the same number of features for \gendesc and GBT feature selection.
We observe that, in general, stability tends to increase as dataset size increases.
Thus, universal features significantly improve stability for datasets with few samples, without reducing predictive performance.
Details are provided in SI Tables \ref{sitab:universal_summary}, \ref{sitab:significance}, and \ref{sitab:significance}, and visualizations are provided in \fig{fig2}c and \sifig{fig2_si}.

\begin{table}
  \centering
  \caption{
    \textbf{Summary of universal features selected by \gendesc:}
    Stability is bounded by $[0,1]$, and higher is better.
    $n = 100$ random train/validation/test splits.
  }
  \label{sitab:universal_summary}
  \begin{tabular}{llccccc}
    \toprule
    \textbf{Type}      & \textbf{Stability} & \textbf{Variance} & \textbf{95\% Confidence Interval} & \textbf{\# Features} & \textbf{Feature Penalty} \\
    \midrule
    Universal Features & 0.3577             & 0.0002            & [0.3287,0.3867]                   & 6.020                & 2.5                      \\
    \bottomrule
  \end{tabular}
\end{table}

\begin{table}
  \centering
  \caption{
    \textbf{Stability of single-task GBT feature selection:}
    Computed for all molecule types and properties in the \all category.
    SM, Prot, and NP stand for ``Small Molecule,'' ``Protein,'' and ``Nanoparticle,'' respectively.
    Stability is bounded by $[0,1]$, and higher is better.
    $n=100$ random train/validation/test splits.
    ``\# Samples" indicates the number of samples available per dataset.
  }
  \label{sitab:task_summary}
  \begin{tabular}{llcccccc}
    \toprule
    \textbf{Type} & \textbf{Property} & \textbf{Stability} & \textbf{Variance} & \textbf{95\% CI}   & \textbf{\# Features} & \textbf{\# Samples} & \textbf{Feature Penalty} \\
    \midrule
    SM            & $\log p$          & 0.6403             & 0.0001            & $[0.6202, 0.6603]$ & 7.7200               & 11,079              & 50                       \\
    Prot          & Solubility        & 0.3691             & 0.0002            & $[0.4312, 0.3970]$ & 6.290                & 2,149               & 10                       \\
    SM            & Boiling           & 0.3003             & 0.0002            & $[0.2742, 0.3264]$ & 7.5000               & 1,185               & 10                       \\
    SM            & $\log h$          & 0.2455             & 0.0001            & $[0.2274, 0.2637]$ & 6.4200               & 777                 & 7.5                      \\
    SM            & Melting           & 0.2148             & 0.0001            & $[0.1916, 0.2381]$ & 6.5300               & 2,143               & 20                       \\
    NP            & $\log p$          & 0.1754             & 0.0002            & $[0.1477, 0.2031]$ & 4.800                & 147                 & 2.5                      \\
    NP            & Zeta Pot.         & 0.1429             & 0.0001            & $[0.1257, 0.1601]$ & 7.0300               & 206                 & 2.5                      \\
    \midrule
    \textbf{Mean} &                   & 0.2983             &                   &                    & 6.6129               &                                                \\
    \textbf{Std}  &                   & 0.1688             &                   &                    & 0.967                &                                                \\
    \bottomrule
  \end{tabular}
\end{table}

\begin{table}
  \centering
  \caption{
    \textbf{Stability of single-task GBT feature selection:}
    Computed for all molecule types and properties in the \all category.
    Stability is bounded by $[0,1]$, and higher is better.
    $n=100$ random train/validation/test splits.
    Cohen's $d$ calculated using stability and variance from Supplemental Tables \ref{sitab:universal_summary} and \ref{sitab:task_summary}.
    $p$-value computed as described in the methodology.
    Note that some values are zero within numerical precision (\ie Python returned 0).
  }
  \label{sitab:significance}
  \begin{tabular}{llcc}
    \toprule
    \textbf{Type}  & \textbf{Property} & \textbf{Cohen's $d$} & \textbf{$p$-value}     \\
    \midrule
    Small Molecule & $\log p$          & -22.21               & 0.000                  \\
    Protein        & Solubility        & -0.788               & 0.5775                 \\
    Small Molecule & Boiling Point     & 4.073                & $3.977 \cdot 10^{-03}$ \\
    Small Molecule & $\log h$          & 9.084                & $1.332 \cdot 10^{-10}$ \\
    Small Molecule & Melting Point     & 10.65                & $5.085 \cdot 10^{-14}$ \\
    Nanoparticle   & $\log p$          & 12.60                & 0.000                  \\
    Nanoparticle   & Zeta Potential    & 17.66                & 0.000                  \\
    \bottomrule
  \end{tabular}
\end{table}

\section{Derivation of \gendesc boosting procedure}
\label{sisec:boosting}

We begin by defining our problem statement and the global objective that we want to optimize.
Define $[n] \doteq \{1, \ldots, n\}$, $\odot$ as the Hadamard product, and $|s|$ as the cardinality of a set.
We further let $\square^t$ indicate the $t$th task, $a_i$ indicate index $i$ for vector $\vec{a}$, $a\ll b$ indicate $a$ is much less than $b$, and $\square_U$ and $\square_{\not U}$ indicate $\square$ for universal features and task-specific features, respectively.
Let $\X \subseteq \Rs^d$ be the feature space and $\Y$ be an abstract output space for regression or classification.
Let each task $t \in [T]$ have a dataset $\D^t \doteq \condset{(\x,y)}{\x \in \X, y \in \Y}$.
In our setting, the data is described by a small subset of features, which can be disjointly split into universal and task-specific feature sets.
Let $\|\cdot\|_0$ indicate the $\ell_0$-pseudo-norm, which counts the number of nonzero elements in a vector.
Thus, in our setting, ${P(y \mid \x)} \approx {P(y \mid (\m^t + \m^U) \odot \x)}$ for ${\|\m^t + \m^U\|_0 \ll d}$, and ${\m^U \odot \m^t = \zero}$, where ${\m^t, \m^U \in \{0,1\}^d}$ indicate binary masks for task $t$ and universal features, respectively.
Then, given a family of parametric functions $\phi_{\vtheta} \in \Phi$ such that $\phi_{\vtheta}\colon \X \to \Y$ for $\vtheta \in \Theta$, loss $\ell: \Y \times \Rs \to [0, \infty)$, and regularizer $R: \Theta \to [0, \infty)$, we aim to find
\begin{equation}
  \label{sieq:global}
  \argmin_{\substack{\vtheta^1, \ldots, \vtheta^T \\ \m^1, \ldots, \m^T, \m^U}}
  \left( \sum_{t\in[T]} \left( \sum_{(\x,y) \in \D^t}
  \ell(y, \phi_{\vtheta^t}((\m^t + \m^U) \odot \x)) \right)
  + \lambda^t\|\m^t\|_0 + \alpha R(\vtheta^t) \right) + \lambda^U\|\m^U\|_0 \quad.
\end{equation}
Here, $\alpha, \lambda^t, \lambda^U \in [0,\infty)$ weight the regularization for parameters $\theta^t$, the task-specific feature sparsity of task $t$, and the universal feature sparsity of all tasks.

Now, we specify the function class $\Phi$ over which we optimize the total.
Define $\T$ as the class of all regression trees, with tree $\tau \in \T: \X \to \Rs$.
For the purpose of analysis, we make the assumption that this set is finite~\citep{chapelle2011} \citep{ustimenko2023}.
This holds in practice, where we learn trees of finite depth using finite datasets.
Indeed, given a finite dataset, all depth-limited trees can be grouped into a finite number of sets, where each set contains trees that provide identical outputs when evaluated on that dataset.
For the purposes of optimization, we only need one tree from each set.
Define $\vtau(\x) \doteq [\tau_1(\x) \dots \tau_{|\T|}(\x)]^{\top} \in \Rs^{|\T|}$ for $\tau_i \in \T$ as the vector of ``tree-transformed features''~\citep{chapelle2011}.
Then, we define $\Phi$ as the set of all additive tree ensembles, with $\phi_{\vtheta^t}(\x) \doteq \sum_{i \in [|\T|]} \theta_i^t\tau_i(\x) = \ip{\vtheta^t}{\vtau(\x)}$ for $\vtheta ^t \in \Theta \doteq \Rs^{|\T|}_+$, as $\T$ is closed under negation.
Note that the feature transformation $\vtau(\cdot)$ is the same across all tasks.
Finally, we make the assumption that $\|\vtheta^{t*}\|_0 \leq B$ (\ie the optimal $\vtheta^{t}$ is $B$-sparse)~\citep{xu2014}.
In other words, given a feature vector $\x$, we non-linearly map it to the space of ``tree-transformed features'' as $\vtau(\x)$, and learn a sparse linear predictor in this ``latent space.''
Note that this method admits both regression and classification, depending on the loss selected.
Thus, we aim to optimize over the class of all additive tree ensembles.

To simplify the selection of universal features, we perform sequential selection of universal and task-specific features.
Recall that we aim to select universal features that are important to all tasks, and task-specific features that are important to one or more (but not all) tasks.
Optimization of our global objective (\sieq{global}) is intractable, so we perform greedy optimization.
In this approach, the first stage selects universal features without interference from task-specific features, and the second stage allows each task to independently select task-specific features while utilizing the previously-selected universal features.
We further optimize each stage greedily using a gradient boosting algorithm, which allows for an efficient boosting-style algorithm for approximating the solution to our global objective.

For the first stage, we show that the trees that select only universal features can be greedily optimized using gradient boosting.
Define $\T^U \subseteq \T$ as trees that only use features that are predictive for all tasks (training datasets), where $\vtheta^t_U \in \Rs^{|\T|}_+$ is the weights for trees in $\T^U$.
Then, we first optimize
\begin{equation}
  \argmin_{\vtheta^1_U, \ldots, \vtheta^T_U} \sum_{\substack{t\in [T] \\ (\x,y) \in \D^t}} \ell(
  y, \ip{\vtau(\x)}{\vtheta^t_U})
  +  \lambda^U\|\m^U\|_0 + \alpha \|\vtheta^t_U\|_1
\end{equation}
to select universal features.
Note that there is no need to optimize $\m^U$ directly, since it is determined by $\theta^t_U, t \in [T]$.
The following derivation is provided in \citep{xu2014}, but is reproduced here for completeness.
We write the subgradient of this total loss for a task $t$ with respect to the parameter $\vtheta^t_U$ as
\begin{equation}
  \left(\sum_{(\x,y) \in \D^t} \frac{\partial}{\partial \vtheta^t_U} \ell(
    y, \ip{\vtau(\x)}{\vtheta^t_U}) \right)
  + \frac{\partial}{\partial \vtheta^t_U} \left(\lambda^U\|\m^U\|_0 + \alpha \|\vtheta^t_U\|_1\right) \;.
\end{equation}
We first derive the subgradient of the regularizers.
Let $F$ indicate a $d$-dimensional feature space, and $\mathbbm{1}_{f}^{t}$ that feature $f$ is previously unused but is included in the current tree for task $t$.
Because the selection of a tree $\tau \in \T^t$ adds $\beta$ to one dimension of the feature vector $\vtheta_{U}^t$ at each boosting round, similar to \citep{xu2014}, we can differentiate the universal feature regularization as
\begin{equation}
  \begin{split}
    \frac{\partial}{\partial \theta_{U, j}^t}\left(\lambda^t \|\m^U\|_0 + \alpha \|\vtheta_{U}^t\|_1 \right)
    = \lambda^t \sum_{f\in F} \mathbbm{1}_{f}^{t} + \alpha \beta \quad,
  \end{split}
\end{equation}
where $\tau_j$ is the tree under consideration, indexed by $\theta^t_{U,j}$.
This penalizes both the addition of new features and new trees to the model.
Now, we focus on the gradient of the loss, where we rewrite as
\begin{align}
  \frac{\partial \ell}{\partial \vtheta^t_U} & = \frac{\partial \ell}{\partial \ip{\vtau(\x)}{\vtheta^t_U}}\frac{\partial \ip{\vtau(\x)}{\vtheta^t_U}}{\partial \vtheta^t_U} \;,
\end{align}
using the chain rule.
$\frac{\partial \ip{\vtau(\x)}{\vtheta^t_U}}{\partial \theta^t_{U,j}} = \tau_j(\x)$ by linearity.
Define $g_{U, b}^t(\x) \doteq - \frac{\partial \ell}{\partial \ip{\vtau(\x)}{\vtheta^t_U}}$.
Then, the subgradient of our total loss is
\begin{equation}
  \sum_{(\x,y) \in \D^t} -g_{U, b}^t(\x) \tau_j(\x) + \lambda^t \sum_{f\in F} \mathbbm{1}_{f}^{t} + \alpha \beta \;.
\end{equation}
Here, we assume that $\sum_{(\x,y) \in \D^t}\tau_j(\x)^2$ is constant~\citep{chapelle2011}.
In theory, this can be achieved by normalizing the output of each tree during optimization.
However, because the trees are grown greedily, we do not normalize them in practice.
Noting that $g_{U, b}^t(\x)$ is a constant with respect to $\tau_j$, we rewrite the optimization problem as
\begin{align*}
  j^t_{b+1} & = \argmin_{\tau_j \in \T^U} \sum_{(\x,y) \in \D^t} - g_{U, b}^t(\x)\tau_j(\x)
  +\lambda^t \sum_{f\in F} \mathbbm{1}_{f}^{t} + \alpha \beta                                                               \\
            & = \argmin_{\tau_j \in \T^U} \sum_{(\x,y) \in \D^t} -2g_{U, b}^t(\x)\tau_j(\x)
  + 2(\lambda^t \sum_{f\in F} \mathbbm{1}_{f}^{t} + \alpha \beta)                                                           \\
            & = \argmin_{\tau_j \in \T^U} \sum_{(\x,y) \in \D^t} g_{U, b}^t(\x)^2 -2g_{U, b}^t(\x)\tau_j(\x) + \tau_j(\x)^2
  + 2(\lambda^t \sum_{f\in F} \mathbbm{1}_{f}^{t} + \alpha \beta)                                                           \\
  j^t_{b+1} & = \argmin_{\tau_j \in \T^U} \sum_{(\x,y) \in \D^t} \left( g_{U, b}^t(\x) -  \tau_j(\x) \right)^2
  +\lambda^t \sum_{f\in F} \mathbbm{1}_{f}^{t} + \alpha \beta \;,\numberthis
\end{align*}
where the last line absorbs the constants into the regularization hyperparameters and $j^t_{b+1}$ is the index of the selected tree for task $t$ at the current boosting round.
In the setting where our loss is the squared error, the term $g_{U, b}^t(\x)$ is the residuals of the previous prediction~\citep{friedman1999}.
As in \citep{xu2014}, we note that the first term is also the squared error, an impurity function.
Thus, the loss structure lends itself to optimization via a boosting-style algorithm.
Using a constant step size, this gives us an additive model defined recursively as ${\vtheta_{U, b+1}^t = \vtheta_{U, b}^t + \beta\ve_{j_{b+1}^t}}$ with $\vtheta_{U, 0} = \zero$, where $\vtheta_{U, b}^t$ is the parameter vector at boosting round $b$ and $\ve_j$ is the $j$th canonical basis vector.
As in \cite{xu2014}, we note that each gradient boosting round increases $\|\vtheta^t_U\|_1$ by $\beta$.
Then, after $r$ boosting rounds, we have $\|\vtheta^t_U\|_1 = \beta r$, meaning that $\ell_1$ regularization is equivalent to early stopping~\cite{friedman1999}.
Thus, we drop the explicit $\ell_1$-regularization terms and limit the number of boosting rounds instead.
With this in mind, we consider $\vtheta^t_{U}$ and $\vtheta_{U, r}^t$ to be equivalent.
This ends our restatement of the derivation from \citep{xu2014}.

Next, we show that the trees that use universal and/or task-specific features can be similarly optimized.
We define $\vtheta^t_{\not{U}} \in \Rs^{|\T|}_+$ as the weights for trees in $\T$ which may use either universal or task-specific features, where the final weight vector will be $\vtheta^t \doteq \vtheta^t_U + \vtheta^t_{\not{U}}$.
Given $\vtheta^t_U$, the optimization of task-specific features is independent for each task $t \in [T]$ and can be written as
\begin{equation}
  \begin{split}
    \argmin_{\vtheta^t_{\not{U}}} \sum_{(\x,y) \in \D^t} \ell(
    y, \ip{\vtau(\x)}{\vtheta^t_U + \vtheta^t_{\not{U}}})
    + \lambda^t\|\m^t\|_0 + \alpha \|\vtheta^t_{\not{U}}\|_1
  \end{split}
\end{equation}
where, as above, $\m^t$ does not need to be directly optimized.
We find the optimal $\vtheta_{\not{U}}^t$ similarly to $\vtheta_U^t$.
We meet the $B$-sparse assumption by letting $B$ equal the total number of boosting rounds (both universal and task-specific).
More generally, we can use $B_U$ and $B_{\not{U}}$ to indicate the number of universal and task-specific boosting rounds, where $B = B_U + B_{\not{U}}$

We greedily learn $\vtheta_U^t$ and $\vtheta_{\not U}^t$ by using gradient boosted CART trees with a penalized impurity function.
$\vtheta_{\not U}^t$ is learned using single-task GBT feature selection, which adds a penalty function to the impurity function, as described in \sec{task_splitting}, \eq{xu}.
However, $\vtheta_U^t$ requires a specialized tree-splitting criterion to ensure that all tasks agree on the universal features.
Thus, we use maximin optimization to grow trees for all tasks at once, as described in \sec{universal_splitting}, \eq{maximin}.
Pseudocode for the universal (SI Alg.~\ref{sialg:csc}) and task-specific (SI Alg.~\ref{sialg:tssc}) splitting conditions and the \gendesc boosting algorithm (SI Alg.~\ref{sialg:boosting}) are provided below.

\begin{algorithm}
  \caption{Universal Splitting Condition}
  \label{sialg:csc}
  \begin{algorithmic}[1]
    \State {\bfseries Input:} Node $\eta^t \subseteq \D_t$ for $t \in [T]$, impurity function $i$.
    \State {\bfseries Output:} Split feature $f_* \in F$ and locations $v^1_*,...,v^T_* \in \Rs$.
    \State \Comment{Use impurity decrease $\Delta i^t_b$ for impurity $i$ on task $t$ and boosting round $b$}
    \State \Comment{Select the feature $f$ that maximizes the minimum impurity decrease of any task}
    \State $f_* \doteq \argmax_{f \in F} \min_{t \in [T]} \max_{v^t \in \Rs} \Delta i^t_b(\eta^t, f, v^t)$
    \State \Comment{Find the splits associated with feature $f_*$}
    \For{$t \gets 1$ to $T$}
    \State $v^t_* \doteq \argmax_{v \in \Rs} \Delta i^t_b(\eta^t, f_*, v)$
    \EndFor
    \State \Return $f_*,v^1_*,...,v^T_*$
  \end{algorithmic}
\end{algorithm}

\begin{algorithm}
  \caption{Task-Specific Splitting Condition}
  \label{sialg:tssc}

  \begin{algorithmic}[1]
    \State {\bfseries Input:} Node $\eta^t \subseteq \D_t$, impurity function $i$.
    \State {\bfseries Output:} Split feature $f_* \in F$ and location $v_* \in \Rs$.
    \State \Comment{Use impurity decrease $\Delta i^t_b$ for impurity $i$ on task $t$ and boosting round $b$}
    \State \Comment{Maximize each impurity decrease independently}
    \State $f_*, v_* \doteq \argmax_{f \in F, v \in \Rs} \Delta i^t_b(\eta^t, f, v)$
    \State \Return $f_*,v_*$
  \end{algorithmic}
\end{algorithm}

\begin{algorithm}
  \caption{\gendesc Boosting Algorithm}
  \label{sialg:boosting}
  \begin{algorithmic}[1]
    \State {\bfseries Input:} Training sets $\D^t \doteq \condset{(\x,y)}{\x \in \X, y \in \Y}$ for $t \in [T]$, an impurity function $i$, number of universal trees $B_U$ and task-specific trees $B_{\not{U}}$, learning rate $\beta$.
    \State {\bfseries Output:} Final model $F^t_{B_U+B_{\not{U}}}$ for $t \in [T]$.

    \State \Comment{Initialize each model with a constant value}
    \State $F^t_0(\x) \doteq 0 \; \forall t \in \set{1,..,T}$

    \State \Comment{Select universal features using ``multitask trees''}
    \State Start with $\eta^t = \D^t\; \forall t\in[T]$
    \For{$b \gets 1$ to $B_U$}
    \Repeat
    \State Grow trees $h^1_b, ..., h^T_b$ with shared splitting feature using Algorithm \ref{sialg:csc} and $\eta^1, ..., \eta^T$
    \Until{Stopping condition reached}
    \State $F^t_{b} \doteq F_{b-1}^t(\x) + \beta h^C_b(\x)$
    \EndFor

    \State \Comment{Select task-specific features independently}
    \State Start with $\eta^t = \D^t\; \forall t\in[T]$
    \For{$b \gets 1$ to $B_{\not{U}}$}
    \For{$t \gets 1$ to $T$}
    \Repeat
    \State Grow tree $h^t_{B_U+b}$ independently using Algorithm \ref{sialg:tssc} and $\eta^t$
    \Until{Stopping condition reached}
    \State $F_{B_U+b}^t(\x) \doteq F_{B_U+b-1}^t(\x) + \beta h^t_{B_U+b}(\x)$
    \EndFor
    \EndFor

    \State \Return $F^1_{B_U+B_{\not{U}}},...,F^T_{B_U+B_{\not{U}}}$
  \end{algorithmic}
\end{algorithm}

\end{appendices}
\twocolumn

\bibliographystyle{sn-nature}
\bibliography{ms.bbl}


\end{document}